\documentclass[conference,a4paper]{style/ieeetran}
\IEEEoverridecommandlockouts
\usepackage{amsmath,amsfonts,amssymb}
\usepackage{graphicx}
\usepackage{gensymb}
\usepackage{amsthm}
\usepackage{cite}
\usepackage[ruled,vlined,linesnumbered]{algorithm2e}
\usepackage[utf8]{inputenc}
\usepackage{hyperref}
\usepackage{caption}
\usepackage{subcaption}
\newtheorem*{assumption*}{Assumptions}

\usepackage{flushend}

\def\BibTeX{{\rm B\kern-.05em{\sc i\kern-.025em b}\kern-.08em
    T\kern-.1667em\lower.7ex\hbox{E}\kern-.125emX}}

\begin{document}


\title{Active Mapping of Underwater Litter Using Camera-Sonar Fusion

\thanks{This work was been financially supported from SeaClear2.0, a project that received funding from the European Climate, Infrastructure and Environment Executive Agency (CINEA) under grant agreement No 101093822.}
}

\author{\IEEEauthorblockN{David Re{\c{t}}e}
\IEEEauthorblockA{\textit{Department of Automation} \\
\textit{Technical University of Cluj-Napoca}\\
Romania \\
dvdrete@gmail.com}
\and
\IEEEauthorblockN{Patrick Boros}
\IEEEauthorblockA{\textit{Department of Automation} \\
\textit{Technical University of Cluj-Napoca}\\
Romania \\
patrick.boros@gmail.com}
\and
\IEEEauthorblockN{Lucian Bu{\c{s}}oniu}
\IEEEauthorblockA{\textit{Department of Automation} \\
\textit{Technical University of Cluj-Napoca}\\
Romania \\
\IEEEauthorblockA{\textit{Corresponding Member of the Romanian Academy}}
Bucharest, Romania \\
lucian.busoniu@aut.utcluj.ro}
}


\maketitle
\thispagestyle{empty}
\pagestyle{empty}

\begin{abstract}
Marine litter is a growing threat to the underwater ecosystem, driving demand for autonomous survey methods that can locate debris efficiently over large areas. Existing survey methods typically follow predefined paths or operate with a single sensing modality, typically a camera (with image quality suffering in poor-visibility conditions) or sonar (usually noisy and low-resolution). We present an active mapping framework in which a forward-looking sonar and a camera both feed into a shared Bayesian occupancy map, and an optimization problem is solved at each step to decide on the next best view. Candidate viewpoints are scored by a two-term utility that balances exploration of uncertain regions via voxel entropy against exploitation of likely objects. Each sensor is characterized by range- and bearing-dependent detection and false-alarm probability tables determined from data. We evaluate the approach in a realistic  underwater simulator, demonstrating that active mapping finds objects faster than a lawnmower coverage pattern, and that the dual-sensor approach works better than using either of the individual sensors.


\end{abstract}


\section{Introduction}\label{sec:intro}

Underwater robots face significant sensing limitations. Cameras suffer from poor optical visibility due to turbidity and inconsistent lighting. Sonars do not need light and are less affected by turbidity, but are often noisier and have poorer resolution than cameras; higher-frequency sonars improve these aspects somewhat at the expense of a narrower field of view, shorter range, and often prohibitive financial costs. These sensing constraints are especially problematic for marine-litter surveys, as debris is often small and/or visually ambiguous. Furthermore, following preplanned paths to search for litter can be a waste of time because it often leads to re-observing already confirmed areas while failing to resolve regions that require additional observations. This motivates active litter mapping, where an underwater robot adaptively chooses its next move so as to concentrate sensing effort on the most informative areas of the seafloor. These challenges are all encountered e.g.~in the SeaClear and SeaClear2.0 projects~\cite{seaclear2}, which focus on a multi-robot system for detection, mapping, and collection of marine litter.

Motivated by the limitations of individual sensors and the need for active mapping, we propose in this paper a method that maintains a litter occupancy map built using Bayesian updates that fuse camera and sonar detections, and selects the next-best-view waypoint that maximizes mapping progress. The maximization objective balances uncertainty reduction (exploration) against confirmation of likely litter (exploitation). In the realistic MARUS underwater simulator, we experimentally show that an uncrewed underwater vehicle (UUV) using our active mapping method finds litter faster than when it uses a uniform-coverage lawnmower pattern; and that fusing the two sensors works better than using either of them individually.

In related work, information-driven planning for robotic mapping and exploration has been extensively studied for terrestrial, aerial, and marine robots, typically using information objectives like entropy reduction. E.g., reference \cite{related3} formulates active 3D mapping as maximization of mutual information over a candidate set of trajectories that combines global shortest-path plans with local motion primitives, followed by gradient-based trajectory optimization that refines the selected trajectory in continuous control space. At a broader level, the survey \cite{related1} categorizes information-driven planners along two axes: the map representation (Gaussian processes for continuous spatial fields vs.\ occupancy grids for structural mapping) and the planning horizon (myopic single-step vs.\ non-myopic multi-step). Our work falls within the occupancy-grid, myopic-planner branch. Differently from most classical methods, which seek to reduce overall map uncertainty -- via e.g.~mutual information, map entropy, or unseen voxel count -- our two-term objective explicitly separates exploration (entropy of uncertain seafloor voxels) from exploitation (count of voxels that belong to likely but unconfirmed objects). This better reflects our litter-search task where confirming litter matters more than obtaining a uniformly low uncertainty across the entire map. 

Methods that balance exploitation and exploration include e.g.\cite{related5}, which proposes an informative path planner for an aerial drone that combines  Gaussian-process maps with an evolutionary trajectory optimizer. The planner reduces overall map uncertainty while using confidence-based level sets to prioritize regions of interest. This line of work is extended in \cite{related6}  with a deep reinforcement learning approach, achieving much faster replanning than optimization. Also related is the field of multitarget search and tracking \cite{related8, related9, related10}, where a very different type of target representation is used, based on intensity functions. Most such approaches focus on tracking targets without exploration, but others do include exploration terms \cite{related7}.

Considering now specifically marine robotics, \cite{related4} addresses tracking of freely drifting surface targets with an ASV using a single stereo camera, by combining a dynamic occupancy grid with a spatiotemporal prediction neural network that estimates future target positions under wind-driven drift, feeding a two-term utility that balances entropy reduction with a redetection reward. In our problem, litter is static on the seafloor. Active viewpoint planning with single acoustic sensors for classifying already detected targets has been studied in \cite{related11, related12}. 

Our key novelty with respect to the work above is that we close the loop between dual, camera-sonar sensor fusion and an active search algorithm for underwater target discovery. The closest related work that uses camera-sonar fusion is in underwater SLAM \cite{related13}, where the robot's pose is not known, so the algorithm focuses heavily on using the map to reduce pose uncertainty, rather than to search for specific targets.

Next, Section \ref{sec:backgd} outlines occupancy grids and neural networks for computer vision, while Section \ref{sec:method} details our method. Experimental results are given in Section \ref{sec:results}, and Section \ref{sec:con} concludes the paper. 
\section{Background}\label{sec:backgd}

\subsection{Occupancy-grid mapping}
We represent the seafloor map as a 3D occupancy grid, consisting of voxels that we index by an integer $i$ for simplicity of notation. Each voxel is associated with a Bernoulli distribution with probability $b_i \in [0,1]$ of being occupied. The real map is $m_i \in \{0, 1\}$, where $m_i=1$ means that voxel $i$ is occupied. Bayesian updates of the voxel grid are performed based on uncertain sensor measurements $z_i$ about $m_i$ for all voxels $i$ in the field of view of the sensor, starting from some prior initial values, which may be taken $0.5$ if we do not have any information in advance:
\begin{equation}\label{eq:bayesupdates}
b^{\prime}_i=\frac{P(z_i \mid m_i=1)\cdot b_i}{P(z_i \mid m_i=0)(1-b_i)+P(z_i \mid m_i=1) \cdot b_i}
\end{equation}
where the probabilities $P(z_i \mid m_i)$ describe sensing uncertainty.

It will be useful to compute the entropy $h_i$ of each voxel $i$:
\begin{equation}
h_i= -\left(b_i\cdot \log(b_i) + (1 - b_i) \cdot \log(1-b_i)\right)
\end{equation}
which is a measure of the uncertainty with which the state of this voxel is known.

The specific implementation that we use is the octree-based Octomap \cite{octomap}, which provides an efficient multi-resolution representation of the 3D environment. 

\subsection{Deep learning for computer vision}
Deep learning methods are widely used for computer vision tasks such as image classification, object detection and segmentation. Convolutional neural networks learn hierarchical feature representations from raw sensor data, enabling extraction of semantic and spatial information \cite{lecun1998gradient}. Image classification assigns a single semantic label to an image, providing scene understanding but not spatial localization \cite{krizhevsky2012imagenet}. Object detection extends this by identifying and localizing multiple objects within the scene allowing for variable object counts \cite{ren2015faster}. Segmentation methods further extend spatial understanding by operating at the pixel level \cite{long2015fully}. While semantic segmentation assigns class labels to pixels, instance segmentation separates individual object instances, which is useful in robotic perception scenarios. In this paper, we use the Mask R-CNN model for instance segmentation, leveraging its two-stage detection architecture, which first generates region proposals for candidate objects and then performs class-specific mask prediction \cite{he2017mask}. 


\section{Active dual-sensor underwater mapping}\label{sec:method}


All experiments reported in this paper are performed in the MARUS underwater simulation environment \cite{marus_simulator}, a Unity-based simulator developed for marine robotics. The simulator provides realistic sensor image rendering and vehicle dynamics; specifically, we use the MiniTortuga UUV \cite{subseatech_minitortuga}, which is equipped with a forward-looking camera and sonar. The simulated scene consists of a static seafloor environment containing multiple objects (cubes and spheres) representing seafloor litter, see Figure~\ref{fig:detection_outputs} below for some views of the scene.

Next, we describe the components of our active search framework: instance segmentation in Section~\ref{sub:object_segmentation}, sensor models in Section~\ref{sub:sensormodels}, map updates in Section~\ref{sub:map_updates}, and the active search planner Section~\ref{sub:active_search}.

\subsection{Instance segmentation from sonar and camera images}\label{sub:object_segmentation}

To perform instance segmentation (including classification) of litter objects, we choose Mask R-CNN with a ResNet-50 backbone and Feature Pyramid Network \cite{he2017mask}, initialized from COCO-pretrained weights \cite{lin2014coco}. This model is selected as it is well-established and stable. Separate models are trained for the sonar and RGB modalities, with no architectural modifications. 

To train the two models, we generated synthetic datasets for both sonar and camera sensors in the MARUS simulator, and used standard training/validation/test data splits. For the sonar dataset, a custom Sonar3D sensor was configured to render forward-looking sonar images. Each dataset contains frames with different object configurations, including scenes with no objects, with a single object, and with multiple objects from both the cube and sphere classes. This variability is essential for characterizing missed detections and false results, and ensures that the network produces meaningful outputs across a broad range of scenes. After constructing the images, we determined camera or sonar hit points belonging to each object instance, and computed their convex hull to generate polygonal object outlines representing ground-truth segmentation masks. The resulting images were stored together with their corresponding ground-truth annotations, including the object class  (cube or sphere) and the polygon vertex coordinates. 

\begin{figure}[!t]
  \centering
  \begin{subfigure}[t]{0.49\linewidth}
    \centering
    \includegraphics[width=\linewidth]{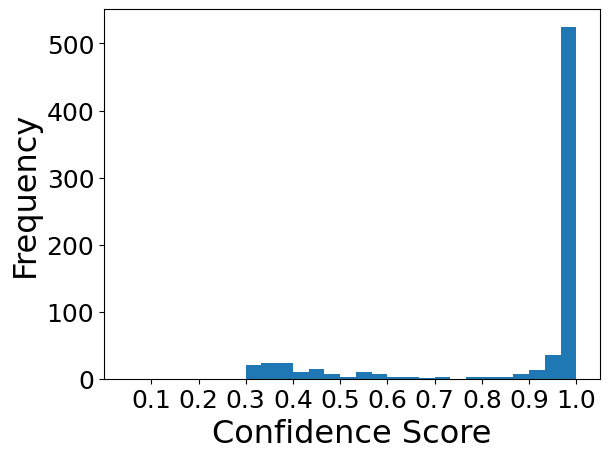}
    \caption{RGB camera}
    \label{fig:confidence_histogram_rgb}
  \end{subfigure}
  \hfill
  \begin{subfigure}[t]{0.49\linewidth}
    \centering
    \includegraphics[width=\linewidth]{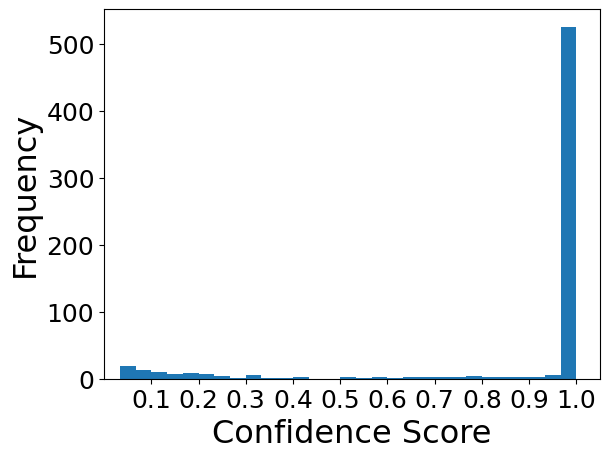}
    \caption{Sonar}
    \label{fig:confidence_histogram_sonar}
  \end{subfigure}
  \caption{Confidence score distributions for the camera and sonar networks evaluated on test data.}
  \label{fig:confidence_histograms}
\end{figure}

\begin{figure}[!t]
  \centering
  \begin{subfigure}[t]{0.49\linewidth}
    \centering
    \includegraphics[width=\linewidth]{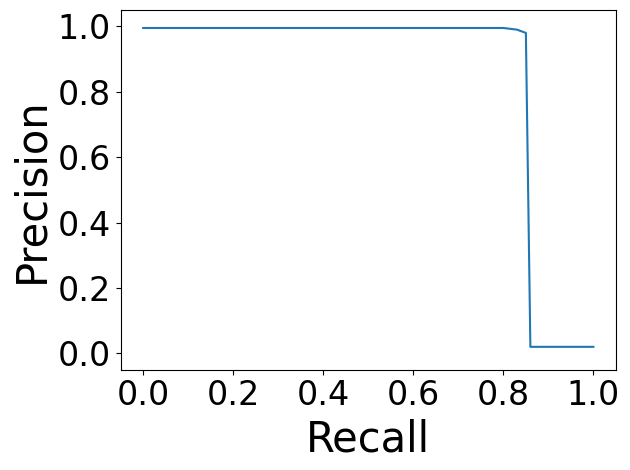}
    \caption{RGB camera}
    \label{fig:pr_curve_rgb}
  \end{subfigure}
  \hfill
  \begin{subfigure}[t]{0.49\linewidth}
    \centering
    \includegraphics[width=\linewidth]{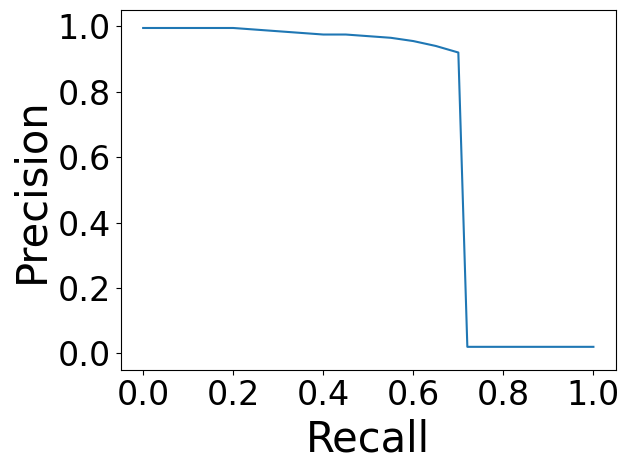}
    \caption{Sonar}
    \label{fig:pr_curve_sonar}
  \end{subfigure}
  \caption{Precision-recall curves for the camera and sonar networks evaluated on test data.}
  \label{fig:pr_curves}
\end{figure}
Results from test data shows that confidence is highly concentrated near one, as illustrated by the confidence score distributions in Figure~\ref{fig:confidence_histograms}.
To understand Figure~\ref{fig:pr_curves}, note first that precision measures the ratio of predicted detections that are correct: $\frac{TP}{TP + FP}$, where $TP$ and $FP$ respectively count true and false positives. Recall measures the ratio of correctly detected objects to the total number of ground-truth objects:
$\frac{TP}{TP + FN}$, where $FN$ is the number of false negatives. The precision-recall curve in Figure~\ref{fig:pr_curves} shows the trade-off between precision and recall as the detection confidence threshold varies. The behavior is near-ideal: for most of the recall range, the precision value remains close to the maximum, while getting true positive detections. This behavior shows a strong score difference between true objects and background clutter, with false positives introduced only when the confidence threshold is driven very low to achieve very large recall.

\begin{figure}[htbp]
  \centering
  \begin{subfigure}[t]{0.49\linewidth}
    \centering
    \includegraphics[width=\linewidth]{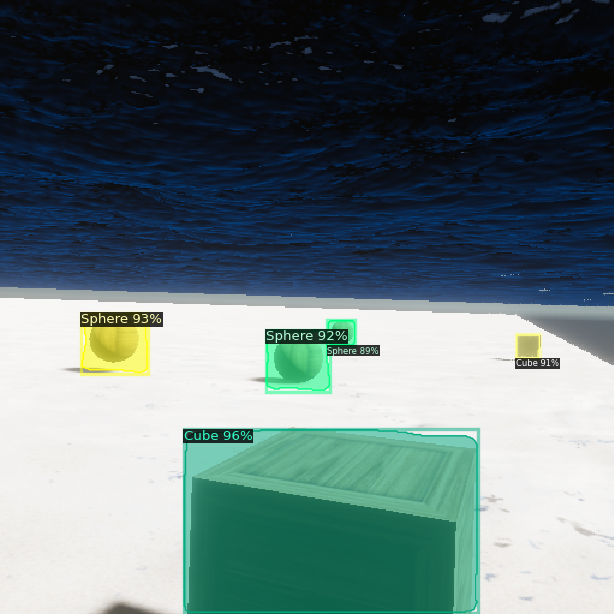}
    \caption{RGB camera}
    \label{fig:detection_outputs_rgb}
  \end{subfigure}
  \hfill
  \begin{subfigure}[t]{0.49\linewidth}
    \centering
    \includegraphics[width=\linewidth]{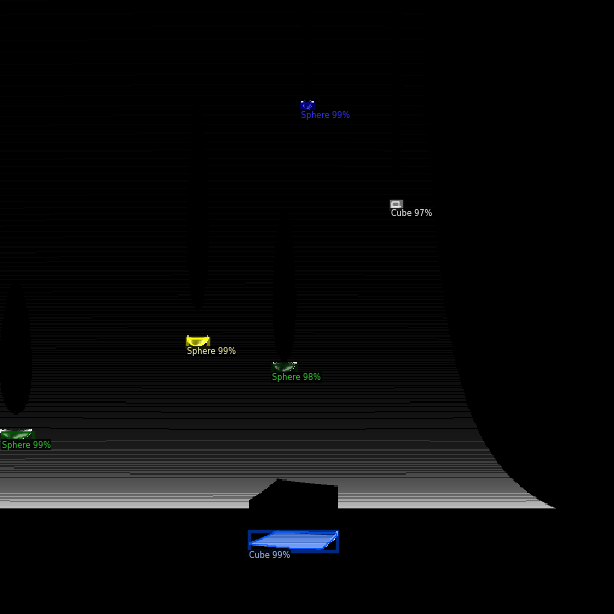}
    \caption{Sonar}
    \label{fig:detection_outputs_sonar}
  \end{subfigure}
  \caption{Instance segmentation outputs for the camera and sonar networks evaluated on test data.}
  \label{fig:detection_outputs}
\end{figure}

Figure~\ref{fig:detection_outputs} illustrates image and sonar detections. Note that bounding boxes are shown for identified classes, but behind the boxes, more precise segmentation masks are visible.

This very good performance is expected given our simulated problem, but is unrealistic for real underwater sensing, where environmental effects and sensor imperfections introduce variability. To bridge the gap between this idealized behavior and realistic sensing uncertainty, we used a temperature scaling parameter applied at inference time to degrade the confidence. The segmentation model generates for each pixel logits $\ell_c$ that represent relative confidence for each class $c$, and the scalar temperature parameter $T$ is applied to the softmax operation to regulate the sharpness of the resulting confidence distribution across classes $c$:
\begin{equation}
P_c =
\frac{\exp\!\left(\ell_c / T\right)}
{\sum_{c'} \exp\!\left(\ell_{c'} / T\right)}
\label{eq:softmax_temp}
\end{equation}
%
Increasing the temperature parameter value makes the confidence scores more uniform and reflects a more realistic, higher uncertainty. 

\subsection{Sensor models}\label{sub:sensormodels}

Let $q$ denote the UUV pose. The pose of each sensor $s \in \{\text{camera}, \text{sonar}\}$ is obtained by a fixed rigid-body transform $q_s = T_s\, q$, where $T_s$ encodes the known mounting offset. To determine measurements $z_i$ for each voxel $i$, a ray is cast from the sensor origin through the center of the voxel. If that ray intersects the image plane inside a segmentation mask for an object, then $z_i=1$; otherwise, i.e.~if the intersection is outside all the segmentation masks, then $z_i=0$. 



We assume that the UUV pose $q$ and therefore the sensor poses $q_s$ are sufficiently well known to disregard their uncertainty, and model sensing uncertainty as follows:
\begin{equation}
\begin{aligned}
P_s(z_i=1 \mid m_i=1) & = \hat t_s(q_s, c_i)\\
P_s(z_i=1 \mid m_i=0) & = \hat f_s(q_s, c_i)
\end{aligned}
\end{equation}
where of course $P_s(z_i=0 \mid m_i) = 1-P_s(z_i=1 \mid m_i)$. Here, $\hat t_s$ and $\hat f_s$ are (approximate) probabilities of a true positive and of a false positive (false alarm), respectively, represented as functions of the sensor pose $q_s$ and the center $c_i$ of voxel $i$. Since our UUV stays horizontal and maintains the same altitude above the seafloor, in practice we will need only a relative pose represented by a scalar range $r(q_s, c_i)$ and bearing $\theta(q_s, c_i)$.



We generated two types of simulated datasets in MARUS to learn the geometry-dependent sensor models for both the sonar and camera. We have target-present runs, used to estimate the true-positive model $\hat t_s$; and empty-scene experiments, in which the target object is removed and any detection will be labeled as a false alarm, to estimate the false-positive model $\hat f_s$ that captures background clutter and spurious detections. For both sensors, data are collected by executing sweeps in which the UUV follows a motion trajectory designed to sample a wide range of relative viewing geometries. In the target-present experiments, the object is placed at fixed known location on the seafloor, and the UUV performs repeated yaw rotations combined with forward-backward translations or circular  orbits to vary range and bearing with respect to target. The same motion pattern is replayed for the empty scene to ensure that the distribution of measurements is matched between the positive and negative datasets. At each step, the corresponding sensor frame (sonar polar image or RGB camera image) is recorded together with the measurement geometry.

\begin{figure}[!tb]
  \centering
  \includegraphics[width=0.4\textwidth]{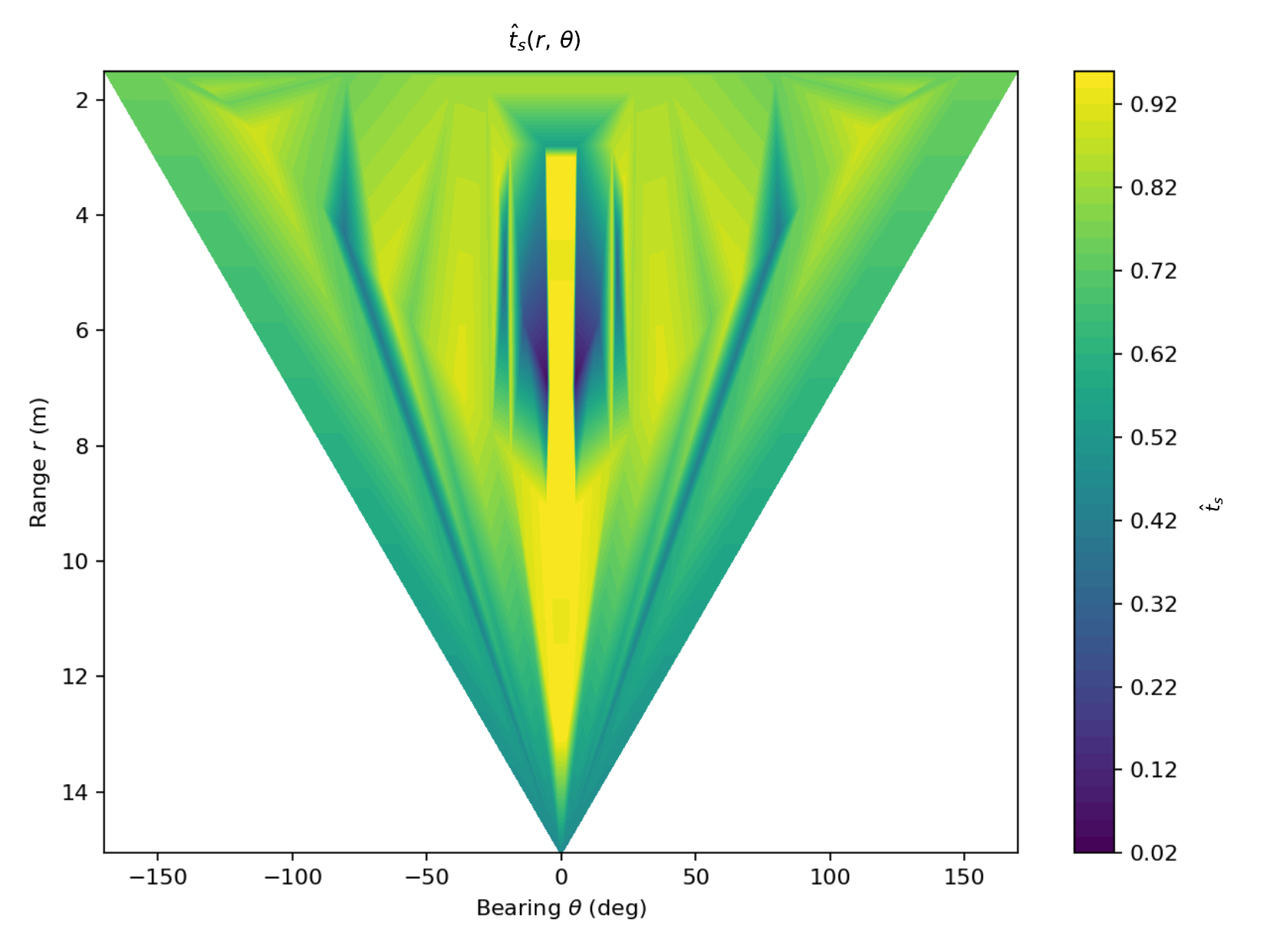}
  \caption{Estimated true-positive probabilities for the camera}
  \label{fig:cam_heatmap}
\end{figure}

\begin{figure}[!tb]
  \centering
  \includegraphics[width=0.4\textwidth]{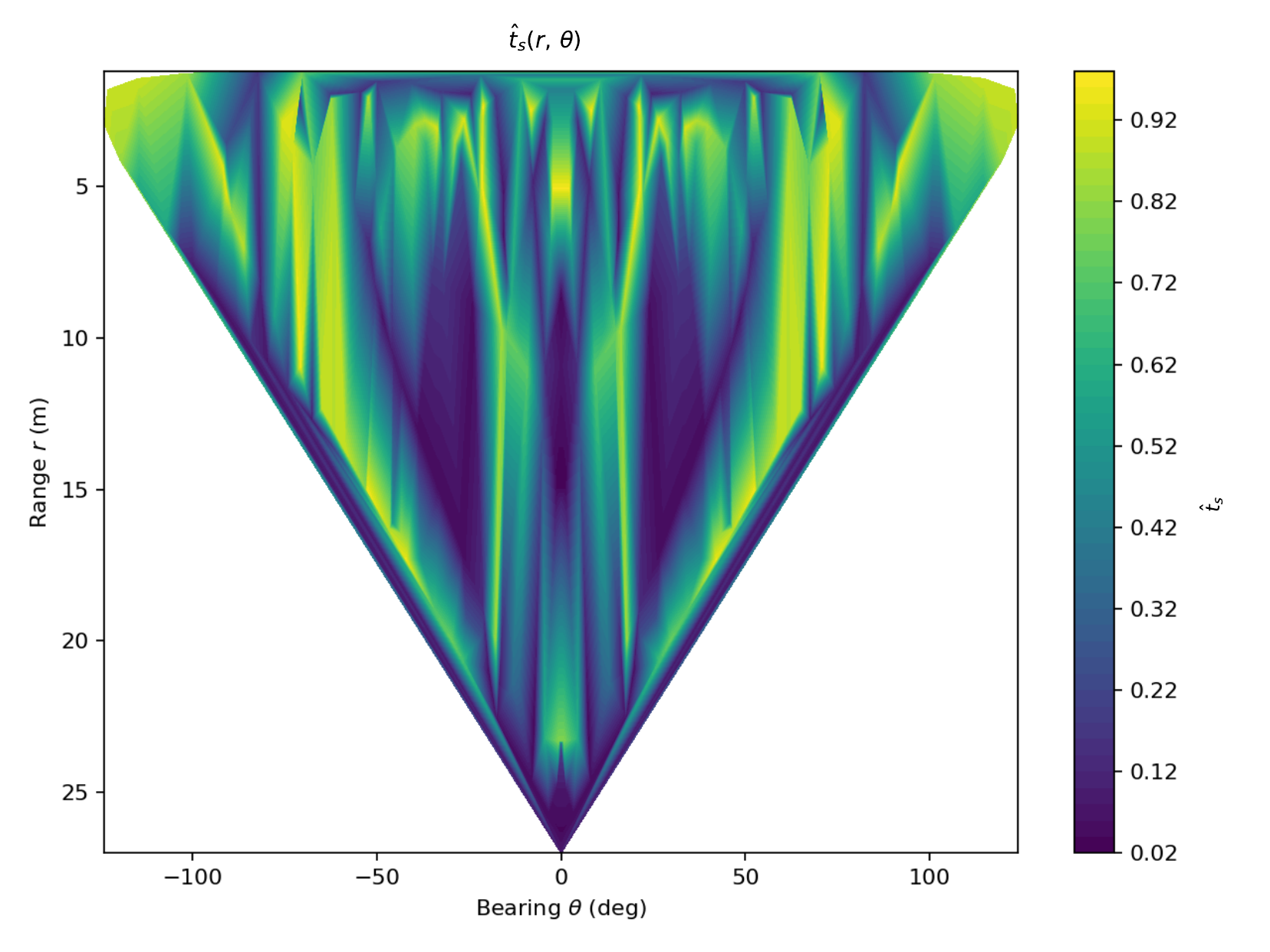}
  \caption{Estimated true-positive probabilities for the sonar}
  \label{fig:sonar_heatmap}
\end{figure}

For each sensor/dataset, the trained Mask R-CNN model is applied independently to every recorded sensor image, producing a binary measurement, indicating whether the target class is detected above a fixed confidence threshold, chosen as 0.8. Each detection is correlated with the corresponding geometry. The measurements are then discretized into bins over these geometric variables, and within each bin the empirical detection frequency is computed as a ratio of positive detections to the total number of samples. This process yields the desired geometry-dependent estimates $\hat t_s$ and $\hat f_s$. The resulting probability tables are visualized in Figure~\ref{fig:cam_heatmap} for the camera and Figure~\ref{fig:sonar_heatmap} for the sonar. The figures use a heatmap representation that encodes how the sensor's reliability varies as a function of viewing geometry, providing a visual representation of regions with high and low detection likelihood.

\subsection{Map and updates}\label{sub:map_updates}

Motivated by the fact that in our real SeaClear system we have access to a bathymetric model of the seafloor, we work here under the same conditions: the seafloor geometry is known as a point cloud representation, which we discretize into voxels of length 0.2m on each side. It is however unknown which voxels correspond to clean seafloor and which to litter objects. The goal is to determine this from sensor readings.

Experiments may use either a single sensor (sonar or camera), or both sensors jointly. For the latter case, because the sonar and camera operate asynchronously at different frequencies, updates may arrive from one or both sensors in a given cycle. Rather than constructing an explicit joint sensor fusion model, we treat the two sensing modalities as conditionally independent given the state of the map. When a single-sensor measurement is received, we simply run \eqref{eq:bayesupdates} while plugging in $\hat t_s$  and $\hat f_s$ for that sensor. When both sensors deliver measurements in the same cycle, the updates are applied sequentially, with the posterior from the sonar serving as the prior for the camera; the same formula \eqref{eq:bayesupdates} is used, but now twice with the two sensor models.


\subsection{Active search}\label{sub:active_search}

Classically, the search for objects is done in a coverage pattern such as a lawnmower or a spiral that investigates the relevant area uniformly. Since searching with the robot costs time and energy, our objective is however to find objects faster than a coverage pattern, by focusing early on promising regions. To this end, we define a set of waypoints $W$ and at each active search step $k$, we choose an optimal next waypoint by solving by enumeration the following optimization problem:
\begin{equation}\label{eq:objective}
w_{k+1} \in  \mathrm{argmax}_{w \in W} \sum_{i \in \phi(w)} \left(h_i + \alpha I(b_i \in [\underline{b}, \overline{b}]) \right)
\end{equation}

Let us explain the elements of the objective function in turn:
\begin{itemize}
\item $\phi(w)$ denotes the field of view: the set of voxels $i$ whose centers fall within the fields of view of the sensors used. Note that when we use both the camera and the sonar, $\phi$ is the union of the two sensors' fields of view. 
\item The \emph{exploration} term $h_i$ gives priority to voxels with larger entropy, i.e.\ larger uncertainty about whether they contain objects or not. If this were the only component, a coverage-like pattern would be followed to uniformly reduce the entropy over the space. However, we want to prioritize object candidates, so we add the next component. 
\item The \emph{exploitation} term focuses on refining object candidates, and is given by an indicator function $I(b_i \in [\underline{b}, \overline{b}])$, which is $1$ if and only if $b_i$ is in the required interval; otherwise, it is $0$. Lower bound $\underline{b} > 0.5$ sets the probability above which a voxel is considered to be part of an object candidate, while  upper bound $\overline{b} < 1$ ensures the algorithm does not unnecessarily focus on voxels that are already clearly part of an object.
\item Weight $\alpha$ trades off exploration versus exploitation, and is the key tuning parameter of the algorithm. A smaller $\alpha$ focuses more on covering unseen areas, while a larger one places focuses on refining object candidates.
\end{itemize}

Each waypoint $w$ is specified by its coordinates $x, y$ on a plane at a given depth, together with the yaw $\psi$ of the UUV. Once a waypoint $w_{k+1}$ has been chosen, the UUV navigates there at a fixed velocity and samples each sensor used in the experiment at its own frequency, updating the map after each sample as expained in Section~\ref{sub:map_updates} above.

\section{Results}\label{sec:results}







This section presents and discusses the results of our active mapping method from joint camera and sonar observations. The simulated scene covers a rectangle of size $[5,38]\times[0,24]\,\mathrm{m}$  and contains 11 objects grouped into three clusters: four cubes in the upper-left corner of the search area (1st cluster), four spheres in the upper right (2nd cluster), and three cubes in the lower right (3rd cluster). The operating depth is \(z_{\mathrm{op}}=-18\,\mathrm{m}\). The exploitation term in \eqref{eq:objective} uses an interval \([\underline{b}, \overline{b}]=[0.7,0.95]\), and the weight $\alpha$ is set to \(5\). The camera operates at \(15\,\mathrm{Hz}\) with a \(60^\circ\) vertical FOV, and the forward-looking sonar operates at \(20\,\mathrm{Hz}\) with a \(120^\circ \times 60^\circ\) FOV and a maximum range of \(30\,\mathrm{m}\). The UUV travels at a nominal velocity of 0.7 $\mathrm{m/s}$, consistent with typical UUV survey operations.

We perform two different experiments. In the first, using both the camera and sonar, we compare active search against a uniform-coverage lawnmower pattern, to check whether the method indeed finds objects faster. In the second experiment, we just use active search and compare the joint-sensor performance with sonar-only and camera-only performance. An object is considered to have been detected when the occupancy probability of at least one voxel whose center is within the ground-truth volume of the object increases above $\overline{b}$.

\begin{figure}[!htbp]
  \centering
  \includegraphics[width=\columnwidth]{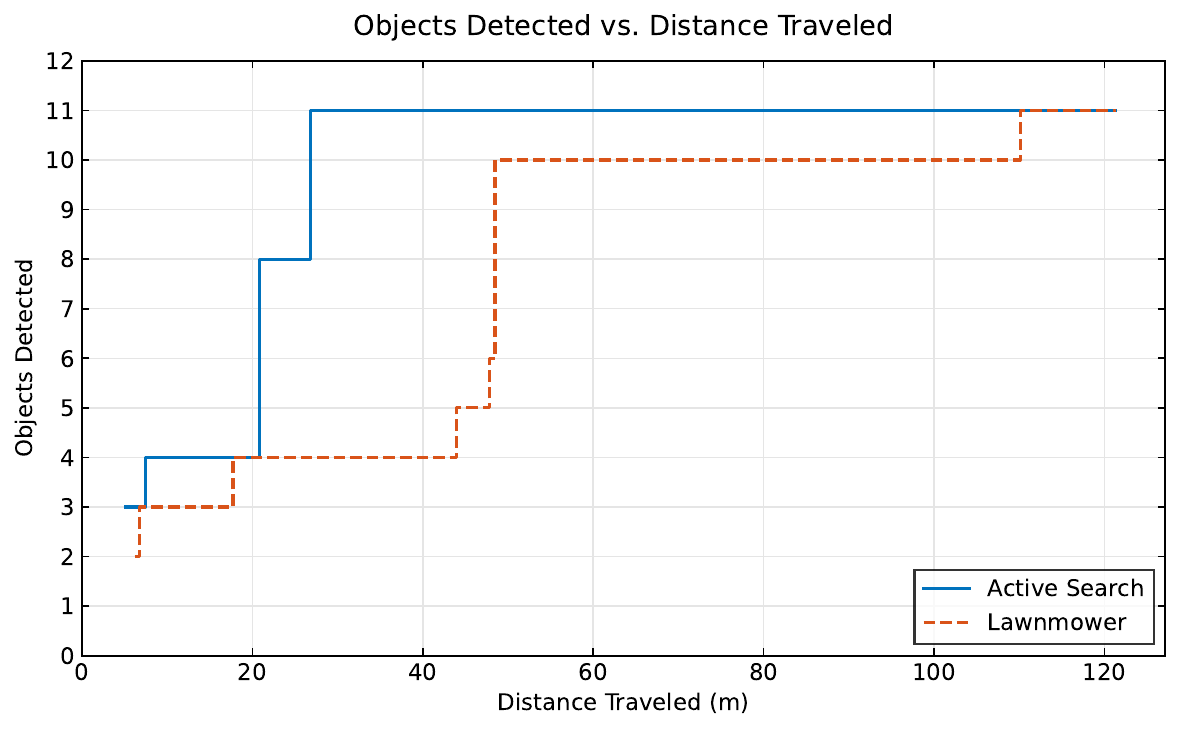}
  \caption{Active mapping vs.~lawnmower mapping using both sensors}
  \label{fig:active_vs_lawnmower}
\end{figure}
\begin{figure}[!htbp]
  \centering
  \includegraphics[width=\columnwidth]{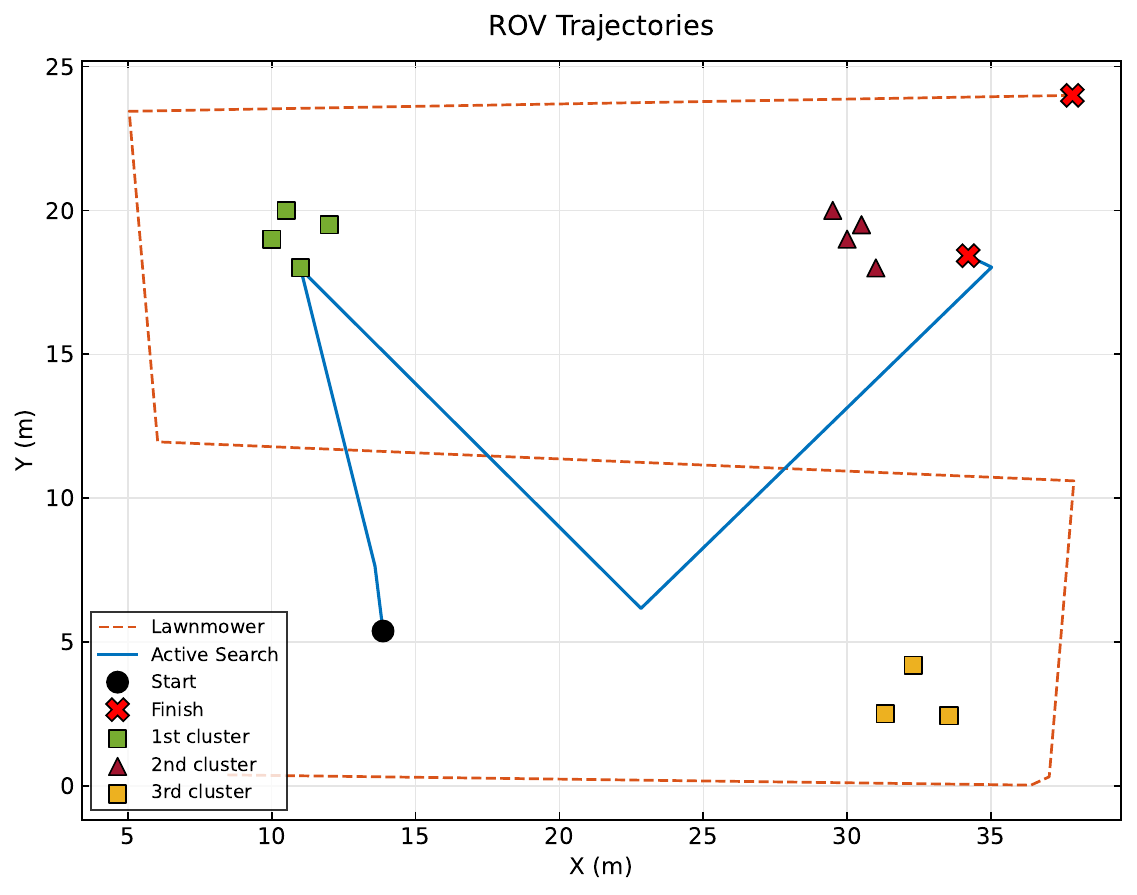}
  \caption{UUV trajectories with the two mapping strategies}
  \label{fig:trajectories}
\end{figure}

For the first experiment, the lawnmower follows a fixed predefined pattern with $12\,\mathrm{m}$ strip spacing. This is a wide spacing that still covers the whole search area in a small number of  passes (specifically, 3 passes). For active search, the grid of waypoints is constructed by uniformly discretizing the search area with grid spacing \(\Delta_x=\Delta_y=6\,\mathrm{m}\) in the plane, and 8 uniformly spaced heading angles at $\Delta_\psi = 45^\circ$ intervals in yaw. Figure~\ref{fig:active_vs_lawnmower} plots the number of detected objects with respect to the distance traveled. The active search strategy is clearly better, since by $27\,\mathrm{m}$ it has already found all 11/11 objects (all 3 clusters), compared with the 4 objects found by the lawnmower. The lawnmower takes $110\,\mathrm{m}$ to find all the objects. Figure \ref{fig:trajectories} illustrates the trajectories performed by the UUV when using the two strategies. Note that active mapping finds the third cluster without having to travel directly on top of it.


\begin{figure}[!htbp]
  \centering
  \includegraphics[width=\columnwidth]{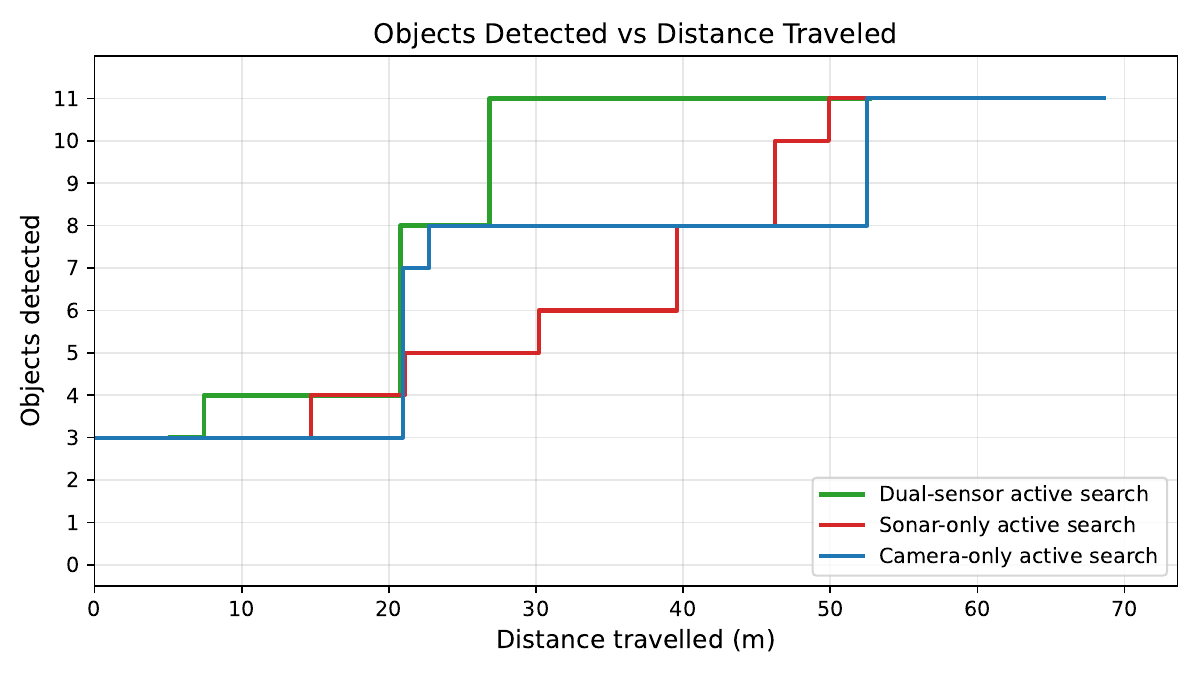}
  \caption{Dual-sensor versus single-sensor performance during active mapping.}
  \label{fig:active_comparison}
\end{figure}

For the second experiment, because the camera has a shorter range than the sonar, to get usable results from camera-only measurements we lowered the grid spacing to \(\Delta_x=\Delta_y=3\,\mathrm{m}\). The rest of the parameters remain the same. The results of this second experiment are shown in Figure~\ref{fig:active_comparison}, where active search with sensor fusion is roughly twice faster in finding all the objects than either the sonar or camera separately. There is no clear winner among the single-sensor setups, further illustrating that the sensors are complementary and should be used together.

\begin{figure}[!htbp]
  \centering
  \includegraphics[width=\columnwidth]{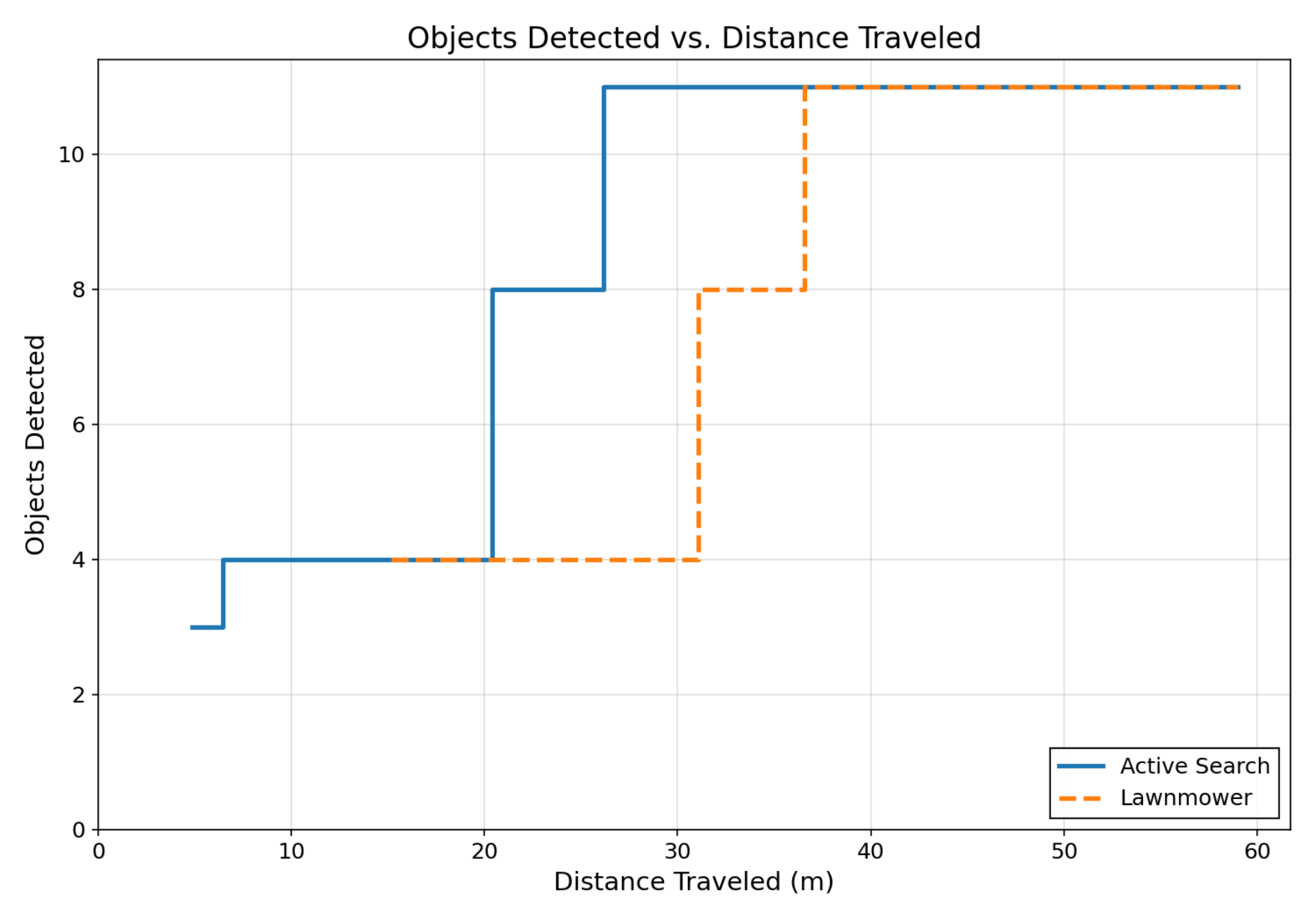}
  \caption{Active mapping vs.~lawnmower mapping for a second scenario}
  \label{fig:scene2_active_vs_lawnmower}
\end{figure}

Finally, we verify the robustness of the method in a different scenario where the object clusters are aligned at the `top' of the scene.
Figure~\ref{fig:scene2_active_vs_lawnmower} shows that active search maintains its advantage.

\section{Conclusions}\label{sec:con}

We presented a method for active, next-best-view search for underwater litter that fuses observations from a forward-looking sonar and a camera into a shared probabilistic map, using sensor models that depend on the range and bearing of the object relative to the sensor. Viewpoints are selected by maximizing an objective function that balances exploration of uncertain regions with exploitation of likely object candidates. Experiments are conducted in a realistic underwater environment simulator, and show that active mapping substantially outperforms predefined fixed-pattern coverage, and that fusing complementary sensor modalities works better than either sensor used separately.


Future work includes developing a joint sensor model that includes couplings between sonar and camera likelihoods; investigating alternate active mapping objectives and non-myopic planners; as well as evaluating the active search pipeline in field trials as a part of the SeaClear2.0 project.

\bibliographystyle{style/IEEEtranS}
\bibliography{aqtr}

\end{document}